\documentclass[10pt,twocolumn,letterpaper]{article}

\usepackage[pagenumbers]{cvpr}

\newcommand{\red}[1]{{\color{red}#1}}

\usepackage{overpic}
\usepackage{pifont}

\usepackage{multirow}

\usepackage{wrapfig}

\usepackage{bbm}
\usepackage{algorithm2e}

\newcommand{\thRows}[1]{\multirow{3}*{#1}}

\newcommand{\tRows}[1]{\multirow{2}*{#1}}

\newcommand{\figref}[1]{Fig.~\ref{#1}}
\newcommand{\tabref}[1]{Tab.~\ref{#1}}
\newcommand{\secref}[1]{Sec.~\ref{#1}}
\newcommand{\algref}[1]{Alg.~\ref{#1}}
\newcommand{\cmark}{\ding{51}}
\newcommand{\xmark}{\ding{55}}

\newcommand{\nameofbase}{partial reconstruction}
\newcommand{\nameofbaseb}{Partial reconstruction}
\newcommand{\myPara}[1]{\noindent\textbf{#1}}
\newcommand{\nameofmethod}[1]{PR-MIM}

\definecolor{hightlightcolor}{gray}{.9}

\definecolor{cvprblue}{rgb}{0.21,0.49,0.74}
\usepackage[pagebackref,breaklinks,colorlinks,allcolors=cvprblue]{hyperref}

\title{\nameofmethod~: Delving Deeper into Partial Reconstruction \\in Masked Image Modeling}

\author{Zhong-Yu Li$^1$ \quad Yunheng Li$^1$ \quad Deng-Ping Fan$^1$ \quad Ming-Ming Cheng$^1$ \\ \\
$^1$VCIP, School of Computer Science, Nankai University \\
}

\begin{document}
\maketitle
\begin{abstract}
    Masked image modeling has achieved great success 
    in learning representations 
    but is limited by the huge computational costs. 
    One cost-saving strategy   
    makes the decoder reconstruct only a subset of masked tokens 
    and throw the others, 
    and we refer to this method as \nameofbase. 
    However, it also degrades the representation quality. 
    Previous methods mitigate this issue 
    by throwing tokens with minimal information 
    using temporal redundancy inaccessible for static images 
    or attention maps that incur extra costs and complexity.
    To address these limitations, 
    we propose a progressive reconstruction strategy and 
    a furthest sampling strategy 
    to reconstruct those thrown tokens in 
    an extremely lightweight way 
    instead of completely abandoning them. 
    This approach involves 
    all masked tokens in supervision to ensure adequate pre-training, 
    while maintaining the cost-reduction benefits of \nameofbase. 
    We validate the effectiveness of the proposed method across various existing frameworks. 
    For example, when throwing 50\% patches, 
    we can achieve lossless performance of the ViT-B/16 
    while saving 28\% FLOPs and 36\% memory usage 
    compared to standard MAE. 
    Our source code will be made publicly available. 
\end{abstract}    

\section{Introduction}

\begin{figure}
  \centering
  \begin{overpic}[width=0.98\linewidth]{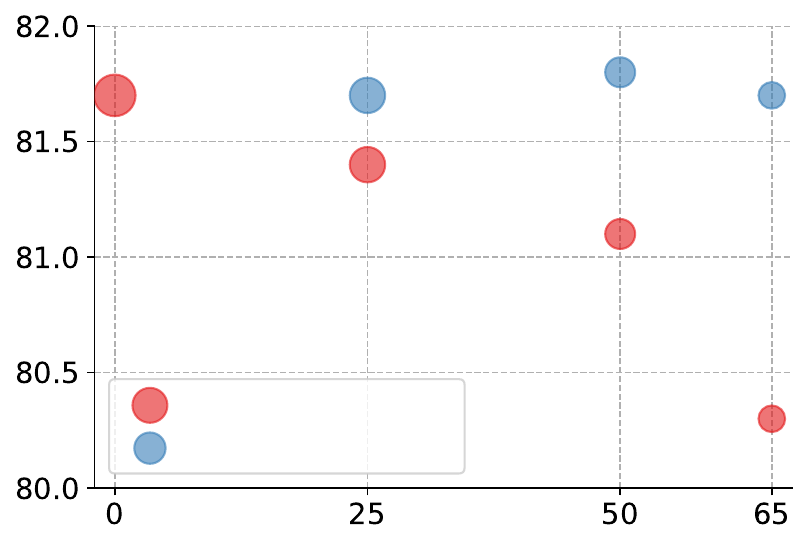}
    \put(22, 10.8){\small Ours}
    \put(22, 16.3){\small \nameofbaseb}
    \put(36, -2){\small token-throwing ratio (\%)}
    \put(-1.5, 23.0){\rotatebox{90}{\small Top-1 accuracy}}
  \end{overpic}
  \caption{Performance on ImageNet-1K. 
  The bubble area is proportional to the training FLOPs.
  A higher throwing ratio indicates more tokens are thrown with lower training costs 
  but leads to greater degradation in plain \nameofbase. }
  \label{fig:decoder_masking_ratio}
\end{figure}
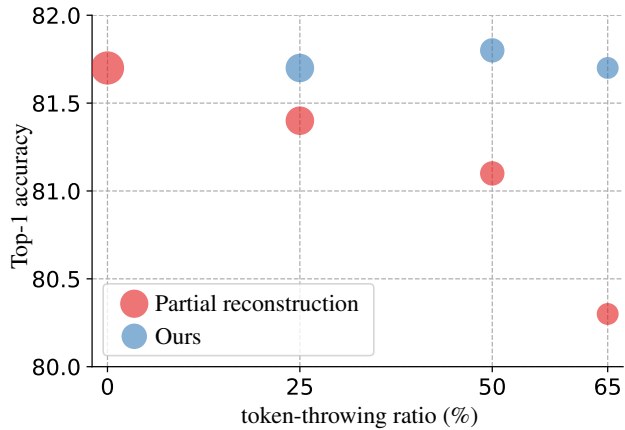

Masked image modeling (MIM), 
which optimizes representations by masking a part of tokens and predicting them, 
has significantly promoted representation learning. 
However, 
it usually demands substantial computational resources, 
\eg, a long pre-training schedule of 1600 epochs~\cite{he2021masked}. 
Recently, 
\nameofbase~\cite{Wang_2023_CVPR,goodhelp,mao2023medical} has been proposed to reduce the costs 
by throwing a subset of masked tokens, 
where these thrown tokens are not reconstructed and involved in the loss calculation, 
thereby reducing computational overhead. 
However, this approach also degrades the quality of learned representations, 
as shown in \figref{fig:decoder_masking_ratio}, 
especially for large models, as shown in our experiments.

Some strategies have been proposed to address the above issue 
but are insufficient for MIM. 
For instance, in masked video modeling~(MVM), 
VideoMAE V2~\cite{Wang_2023_CVPR} leverages temporal redundancy 
to throw tokens that are repeated across adjacent frames, 
effectively mitigating the adverse effects of \nameofbase~in MVM. 
However, the temporal redundancy is inaccessible for static images. 
In MIM, 
\cite{goodhelp,mao2023medical} utilize attention maps 
to throw the tokens with low attention scores. 
However, 
it requires  
running a pre-trained transformer on the complete token sequence and 
consumes storage space, 
which contradicts the goal of \nameofbase~to reduce costs. 
Moreover, 
these methods cannot eliminate the degradation when 
applying \nameofbase~to some MIM frameworks like SimMIM~\cite{Xie_2022_CVPR}. 
Instead, 
we aim to bridge the performance gap while 
preserving the training efficiency of \nameofbase{} and 
apply our method to various MIM frameworks.

\nameofbaseb~throws a part of tokens that can contribute to effective supervision, 
thus reducing the effective epochs and 
the accuracy of gradients estimated in stochastic gradient descent. 
Thus, 
we propose a progressive reconstruction scheme to 
reconstruct the tokens thrown by \nameofbase, 
allowing them to contribute to model supervision. 
After the decoder first reconstructs 
the tokens retained by \nameofbase, 
we use an extremely lightweight spatial aggregation module to reconstruct the thrown tokens 
based on those retained tokens. 
This spatial aggregation 
requires negligible $7.3\cdot 10^{-3}$G floating-point operations per second~(FLOPS) and 
increases almost no costs. 
Experiments 
show that this approach mitigates the performance degradation caused by \nameofbase~and 
yields reasonable reconstruction results. 
Furthermore, 
we obverse an unfavorable situation, 
\ie, 
there may be too few retained tokens 
surrounding a thrown token. 
This phenomenon is not conducive to effectively 
reconstructing this thrown token.
Thus, we propose a furthest sampling strategy, 
which maximizes the spatial dispersion of the tokens retained by \nameofbase, 
to reduce the occurrence of this situation.
Moreover, this operation can be integrated into 
the multi-threaded data augmentation process, 
thus not impeding training speed. 

In this work, 
we integrate the proposed method with different frameworks, 
including MAE~\cite{he2021masked}, SimMIM~\cite{Xie_2022_CVPR}, 
TEC~\cite{tec}, 
GreenMIM~\cite{huang2022green}, LocalMIM~\cite{wang2023masked}, and MFF~\cite{Liu_2023_ICCV}, 
Through experiments with different 
model sizes, training schedules, and token-throwing ratios, 
we demonstrate that the proposed method can 
mitigate the performance degradation caused by \nameofbase, 
as shown in \figref{fig:decoder_masking_ratio}. 
For example, 
compared to MAE, 
when training the ViT-B/16 for 800 epochs and throwing 50\% tokens, 
we achieve consistent performance on various downstream tasks 
with only 72\% GFLOPs, 75\% pre-training time, and 64\% memory usage. 
Further reduction can be achieved by throwing more tokens without 
sacrificing the performance. 

In summary, our main contributions include: 
1) We design the progressive reconstruction and the furthest sampling strategy to 
mitigate the performance degradation caused by \nameofbase~while 
keeping its ability to save costs, 
2) The proposed method can significantly reduce the costs 
without sacrificing performance, 
facilitating the development of efficient representation learning framework, 
and 3) Our method is orthogonal to various methods, 
and we validate its effectiveness on various downstream tasks.

\section{Related works}
\myPara{Self-supervised learning.}
Self-supervised learning~(SSL) learns rich representations without relying on human annotations. 
Early works try different pretext tasks, 
\eg, coloration~\cite{zhang2016colorful,Larsson_2017_CVPR}, 
jigsaw puzzles~\cite{norooziECCV16}, 
rotation~\cite{gidaris2018unsupervised},
autoencoder~\cite{Doersch_2015_ICCV}, 
inpainting~\cite{Pathak_2016_CVPR}, and 
counting~\cite{Noroozi_2017_ICCV}, 
but these methods only achieve unsatisfactory performance. 
The recent success is partly 
due to the rapid development of instance discrimination. 
Assuming that representations 
should be invariant to image transforming, 
these methods~\cite{Wu_2018_CVPR,Zhao_2021_ICCV,Dwibedi_2021_ICCV,Koohpayegani_2021_ICCV,mugs2022SSL,roh2021scrl} pull 
representations extracted from different views of an image closer. 
In this framework, 
different classical methods propose different strategies to avoid model collapse and guarantee the quality of learned representations, 
\eg, contrastive learning with negative samples~\cite{He_2020_CVPR,chen2020simple,Henaff_2021_ICCV,wang2020DenseCL}, 
asymmetric architecture~\cite{Chen_2021_CVPR,byol,Xie_2021_CVPR,DynamicsContrastive}, 
feature decoupling~\cite{DecoupledContrastive,barlow_Twins,ermolov2021whitening}, 
and uniform distribution over channels~\cite{caron2021emerging,caron2020unsupervised}. 
Some works also try different forms of representations in SSL, 
including self-relation~\cite{li2023sere} and 
category assignment~\cite{caron2021emerging,caron2020unsupervised,asano2020self}. 
Due to the increasingly powerful computing devices, 
these methods achieve excellent results on various vision tasks 
but require significant costs.

\myPara{Masked image modeling.} 
Inspired by masked language modeling~\cite{devlin2019bert}, 
masked image modeling~(MIM)~\cite{bao2021beit,he2021masked,li2022semmae,tian2023designing,Woo2023ConvNeXtV2} 
has successfully developed to learn image representations. 
MIM demonstrates superior performance by 
masking a subset of tokens and then 
reconstructing them based on the representations encoded from the unmasked tokens. 
The development of MIM is accompanied by the continuous optimization of reconstruction targets, 
which influences the properties of the learned representations. 
For example, 
many works~\cite{he2021masked,Xie_2022_CVPR,hdm2023,tian2023integrally,Liu_2023_ICCV} use pixel values as the target. 
The hand-designed HOG feature~\cite{Wei_2022_CVPR,1467360} and frequency feature~\cite{xie2023masked}  
have proven to be more effective targets than pixels. 
Many methods further attempt to extract representations from pre-trained models as 
the reconstruction target, 
including the pre-trained tokenizer in~\cite{bao2021beit,dong2022peco,beit3}, 
pre-trained representation encoder in~\cite{tec,wei2022mvp}, 
and siamese network in~\cite{Tao_2023_CVPR,chen_context_2024,Assran_2023_CVPR,10330745,dong2022bootstrapped}. 
Some works~\cite{yi2022masked,Tao_2023_CVPR,zhou2021ibot} also take contrastive representations as the target 
to integrate the advantage of instance discrimination into MIM. 
Unlike instance discrimination, 
which requires multiple views of an image, 
MIM operates on a single view, 
allowing for larger batch size and faster training. 
However, 
the costs are still huge. 
In this work, 
we focus on leveraging \nameofbase~to save costs 
while avoiding the issue that \nameofbase~degrades the quality of representations. 

\begin{figure}[t]
  \begin{overpic}[width=1.0\linewidth]{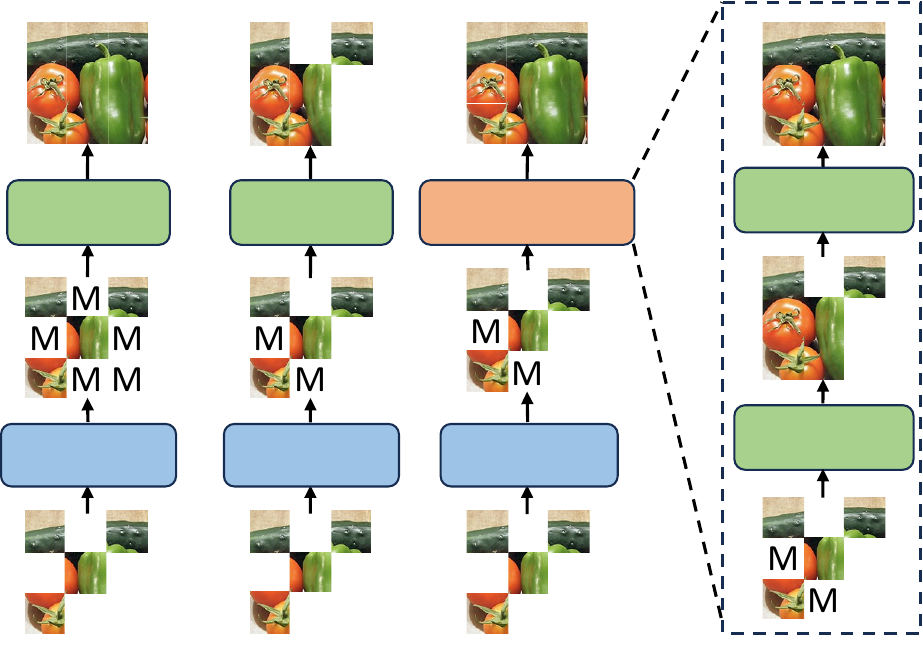}
    \put(03.5, 19.8){\small encoder}
    \put(28.0, 19.8){\small encoder}
    \put(51.2, 19.8){\small encoder}
    \put(03.6, 47.4){\small decoder}
    \put(06.5, 44.4){\scriptsize 5.3G}
    \put(27.5, 47.4){\small decoder}
    \put(30.8, 44.4){\scriptsize 2.6G}
    \put(48, 47.8){\small progressive}
    \put(46.5, 44.6){\small reconstruction}
    \put(69.5, 46.0){\small \red{2.6G}}
    \put(83.0, 23.0){\small decoder}
    \put(85.8, 20.4){\scriptsize \red{2.6G}}
    \put(80.8, 47.2){\small \red{7.3$\cdot 10^{-3}$G}}
    \put(7.2, -3){(a)}
    \put(4.8, -8){\small MAE}
    \put(31.5, -3){(b)}
    \put(18.5, -8){\small \nameofbaseb}
    \put(55.5, -3){(c)}
    \put(54.5, -8){\small Ours}
  \end{overpic}
  \vspace{1pt}
  \caption{Different strategies 
  for masked image modeling and 
  the corresponding FLOPs of the decoder. 
  \nameofbaseb~throws a part of masked tokens $x_t$ and retains the others $x_d$~(marked by $\rm M$), 
  while our progressive reconstruction scheme proposed in 
  \secref{sec:progressive_reconstruction} reconstructs the thrown tokens 
  with minimal computational costs to ensure adequate training.}
  \label{fig:compare}
\end{figure}

\section{Method}

\subsection{Preliminary}

\begin{figure*}[t]
  \begin{overpic}[width=0.98\linewidth]{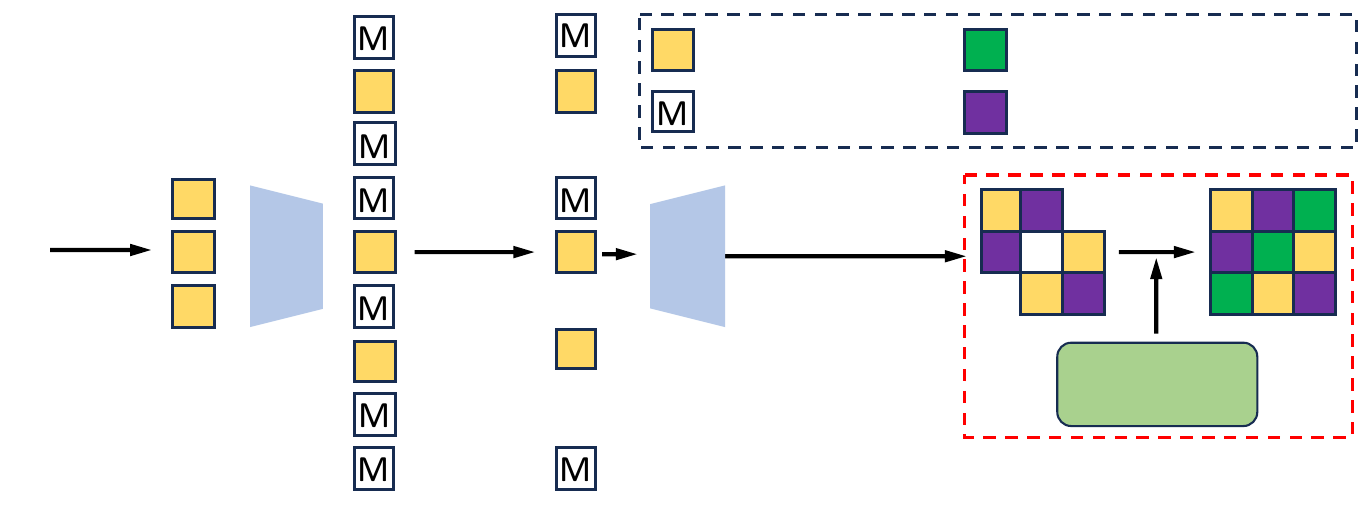}
    \put(0.0, 16.8){\rotatebox{90}{image}}
    \put(3.8, 21.0){masking}
    \put(31.3, 23.5){throwing}
    \put(32.2, 21.0){tokens}
    \put(20.3, 15.7){\rotatebox{90}{encoder}}
    \put(50.0, 15.7){\rotatebox{90}{decoder}}
    \put(82.2, 10.3){spatial}
    \put(80.6, 7.7){aggregation}
    \put(71.8, 3.0){reconstructing the thrown tokens}
    \put(74.5, 0.0){with $7.3\cdot 10^{-3}$ GFLOPs}
    \put(52.5, 33.5){unmasked tokens $x_u$}
    \put(52.5, 29.0){[MASK] tokens}
    \put(75.5, 33.5){thrown masked tokens $x_t$}
    \put(75.5, 29.0){unthrown masked tokens $x_d$}
  \end{overpic}
  \caption{When \nameofbase~throws a subset of masked tokens, 
  the proposed progressive reconstruction scheme 
  reconstructs each masked token with minimal additional costs.}
  \label{fig:framework}
\end{figure*}

\myPara{Masked image modeling.}
Masked image modeling~(MIM) learns representations by 
masking a subset of tokens 
and then reconstructing the masked tokens based on the unmasked tokens. 
MIM has various implementations and 
we take the asymmetric encoder-decoder design in MAE~\cite{he2021masked} as an example to illustrate our method. 
Specifically, following ViT~\cite{dosovitskiy2020vit}, 
an image is first divided into a sequence of regular non-overlapping tokens $x\in \mathbb{R}^{N\times C}$, 
where $N$ is the number of patches and $C$ is the number of dimensions. 
Then, a masking operation randomly masks $N\cdot\rho_e$ tokens 
and feeds the remaining $x_u\in \mathbb{R}^{N\cdot(1-\rho_e)\times C}$
into an encoder to extract representations, 
where $\rho_e$ is the masking ratio and typically set to 75\%~\cite{he2021masked}. 
As shown in \figref{fig:compare} (a), 
the decoder takes representations encoded in $x_u$ and 
a trainable [MASK] token as input, 
where the [MASK] token is copied $N\cdot\rho_e$ times and 
then placed in the positions of masked tokens. 
In the decoder, 
the masked tokens are reconstructed by aggregating representations of 
unmasked tokens $x_u$, 
prompting the encoder to extract rich context information. 
In inference and fine-tuning, 
the encoder is used, 
and the decoder is removed. 

\myPara{\nameofbaseb.}
\nameofbaseb~\cite{Wang_2023_CVPR,goodhelp,mao2023medical} 
has recently been utilized 
to reduce the computational costs of MIM 
by throwing a subset of masked tokens and only reconstructing the remaining ones, 
as shown in \figref{fig:compare} (b). 
For simplicity, 
we define $\rho_d$ as the token-throwing ratio, 
meaning a subset of masked tokens $x_t \in \mathbb{R}^{N\cdot \rho_d\times C}$ are thrown. 
This operation allows the decoder to operate on 1) unmasked tokens $x_u\in \mathbb{R}^{N\cdot(1-\rho_e)\times C}$ 
and 2) masked but unthrown tokens $x_d\in \mathbb{R}^{N\cdot (\rho_e-\rho_d)\times C}$, 
with $N\cdot(1-\rho_d)$ tokens. 
Thus, 
it accelerates pre-training, 
especially for transformer-based architectures where the attention mechanism 
exhibits quadratic complexity to the sequence length. 

\subsection{Progressive reconstruction scheme}
\label{sec:progressive_reconstruction}
While reducing costs, 
\nameofbase~only leverages a subset of tokens to supervise models in each training iteration, 
reducing the number of effective epochs and 
making the mini-batch-level stochastic gradient estimation less accurate. 
In this work, we propose a progressive reconstruction scheme that 
consists of 
a common decoder described in~\cite{he2021masked} and 
an extremely lightweight spatial aggregation module, 
where the former throws a subset of masked tokens and 
the latter reconstructs the thrown tokens in a simple yet effective manner, 
as shown in \figref{fig:compare} (c). 
The framework of our method is demonstrated in 
\figref{fig:framework}. 

\begin{table}[t]
  \setlength{\tabcolsep}{0.8mm}
  \begin{tabular}{lll} \toprule
    Sym. & Dim. & Meaning \\ \midrule
    $N$ & scalar & number of tokens  \\
    $\rho_e$ & scalar & masking ratio \\
    $\rho_d$ & scalar & throwing ratio \\
    \midrule
    $x$ & $N\times C$ & patch tokens, $x=x_u\cup x_d\cup x_t$ \\
    $x_u$ & $N\cdot(1-\rho_e)\times C$ & unmasked tokens \\
    $x_d$ & $N\cdot (\rho_e-\rho_d)\times C$ & masked and unthrown tokens \\
    $x_t$ & $N\cdot \rho_d\times C$ & masked and thrown tokens \\
    \bottomrule
  \end{tabular}
  \caption{Table of symbols, their dimensions, and meaning.}
  \label{tab:Symbols}
\end{table}

\myPara{Spatial aggregation.} 
In the decoder, 
the unthrown masked tokens $x_d$ are reconstructed 
by acquiring context information from the unmasked tokens $x_u$. 
We also reconstruct the thrown masked tokens $x_t$ by aggregating the information from $x_u$ and $x_d$. 
Differently, 
channel-level modeling, \eg, FFN, 
is not required 
because the decoder has mapped the $x_u$ and $x_t$ to the space of reconstruction targets. 
Moreover, 
we aim to design a spatial aggregation module as simple and efficient as possible 
to save computational costs. 
In experiments, 
we show that 
a simple depth-wise convolution is enough to 
reconstruct $x_t$. 

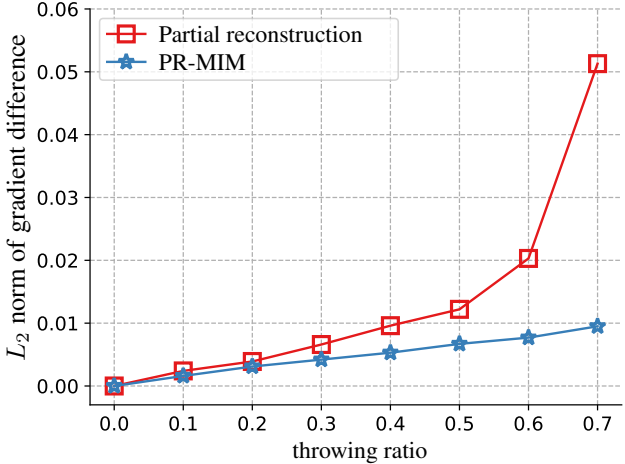
\begin{figure}[t]
  \hspace{5pt}
  \begin{overpic}[width=0.97\columnwidth]{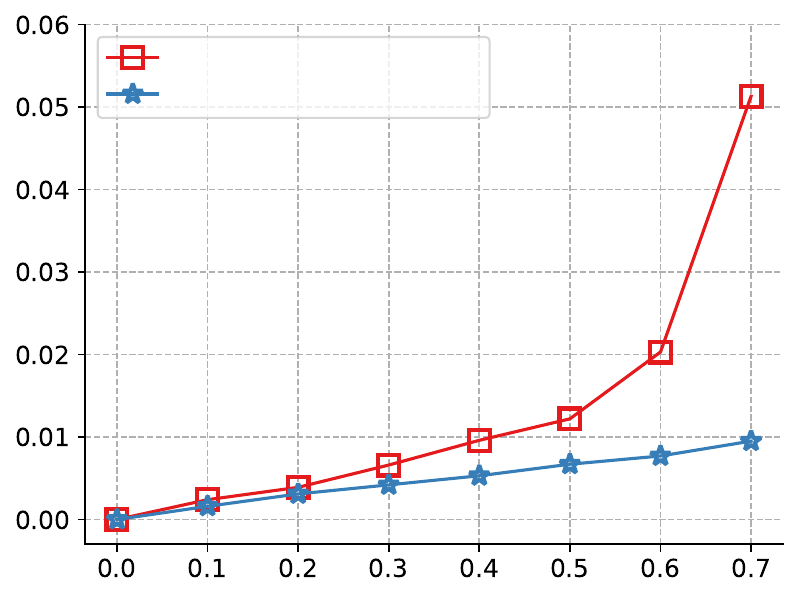}
    \put(44, -2){\small throwing ratio}
    \put(-3, 14){\rotatebox{90}{$L_2$ norm of gradient difference}}
    \put(22, 66.5){\small \nameofbaseb}
    \put(22, 61.8){\small \nameofmethod{}}
  \end{overpic}
  \caption{The $L_2$ norm of gradient difference between different methods and standard MAE.}
  \label{fig:gradient}
\end{figure}

Specifically, 
given $x_u$ and $x_d$ from the decoder output, 
we first fill the thrown tokens $x_t$ with zero values 
and 
rearrange all tokens from sequence to spatial format. 
Then, we apply a $7\times7$ depth-wise convolution to reconstruct 
the thrown tokens 
by aggregating information from $x_u$ and $x_d$ output by the decoder, 
with negligible $7.3\cdot 10^{-3}$GFLOPs. 
Then, 
we predict the pixel values of each masked token using a normalization layer and a linear layer, 
as described in~\cite{he2021masked}. 
In this way, 
involving each masked token in supervision
can ensure adequate training 
and estimate gradients more accurately. 
As shown in \figref{fig:gradient}, 
there is a large deviation between the gradient estimated by \nameofbase~and standard MAE, 
especially when using a large throwing ratio. 
In contrast, our \nameofmethod{}~can significantly alleviate this problem.

\subsection{Furthest sampling} 
\begin{figure}
  \centering
  \begin{overpic}[width=\linewidth]{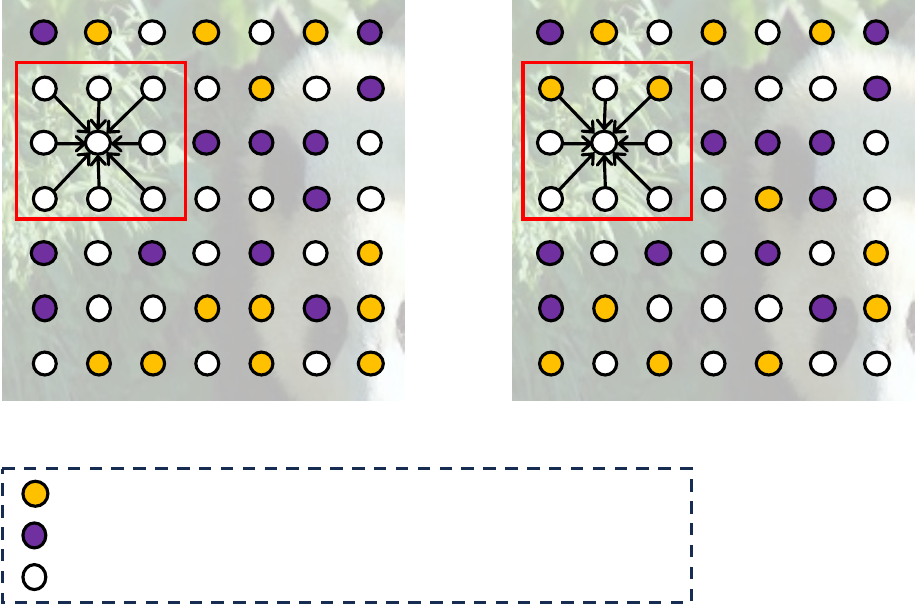}
    \put(8, 18){random sampling}
    \put(64, 18){furthest sampling}
    \put(7, 11){unmasked tokens $x_u$}
    \put(7, 6.4){masked and unthrown tokens $x_d$}
    \put(7, 2){masked and thrown tokens $x_t$}
  \end{overpic}
  \caption{The furthest sampling strategy. 
  The red box and arrows mean that the thrown token at the center 
  is reconstructed by aggregating information from the other tokens 
  within the box.}
  \label{fig:visualization_furthest}
\end{figure}
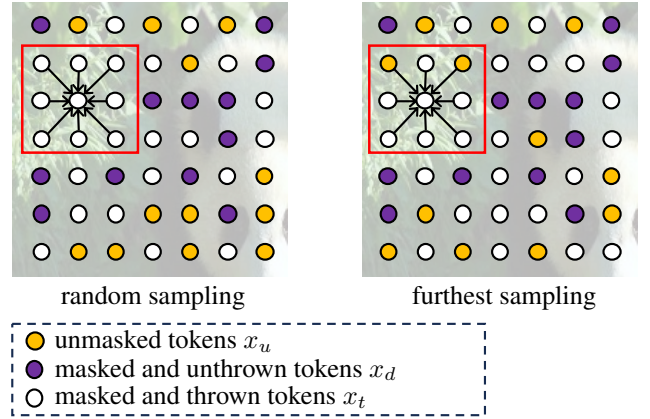

To reconstruct each thrown masked token in $x_t$, 
it is essential to ensure that 
there are enough tokens from $x_u$ and $x_d$ 
within a $7\times 7$ region around each token in $x_t$, 
enabling each token in $x_t$ to aggregate sufficient context information. 
However, 
this requirement is not ensured when randomly throwing tokens. 
Especially when there is no token of $x_u$ and $x_d$ around a token in $x_t$, 
as shown in \figref{fig:visualization_furthest} (left), 
we cannot calculate the meaningful loss on this token. 
A straightforward manner to overcome this issue 
is to enlarge the receptive field of the spatial aggregation module. 
However, 
experiments in \tabref{tab:kernel_size} show that an excessive receptive field is sub-optimal, and 
previous works~\cite{chen2022efficient} have shown that a token is reconstructed by aggregating local information. 
To address this issue, 
we adopt a furthest sampling strategy, 
which selects the unthrown tokens $x_d$ as dispersedly distributed as possible, 
as shown in \figref{fig:visualization_furthest} (right).

\begin{table*}[t]
  \centering
  \setlength{\tabcolsep}{4.0mm}
  \begin{tabular}{ccclccc} 
    Progressive Reconstruction & Furthest Sampling & $\rho_d$ & Top-1 & FLOPs & Memory & Time \\ \midrule
    \xmark & \xmark & 0 & 81.7 & $1.00\times$ &  $1.00\times$ & $1.00\times$ \\
    \midrule
    \xmark & \xmark & 25 & 81.4$_{\color{red} -0.3}$ & \multirow{4}*{$0.85\times$} & \multirow{4}*{$0.81\times$}  & \multirow{4}*{$0.83\times$}\\
    \cmark & \xmark & 25 & 81.7 & \\
    \xmark & \cmark & 25 & 81.6 & \\
    \cmark & \cmark & 25 & {81.7$_{\color{ForestGreen} -0.0}$} & & \\
    \midrule
    \xmark & \xmark & 50 & 81.1$_{\color{red} -0.6}$ & \multirow{4}*{$0.72\times$} & \multirow{4}*{$0.64\times$}  & \multirow{4}*{$0.75\times$} \\
    \cmark & \xmark & 50 & 81.6 & & \\
    \xmark & \cmark & 50 & 81.4 & & \\
    \cmark & \cmark & 50 & {81.8$_{\color{ForestGreen} +0.1}$} & & \\
    \midrule
    \xmark & \xmark & 65 & 80.3$_{\color{red} -1.4}$ & \multirow{4}*{$0.64\times$} & \multirow{4}*{$0.55\times$} & \multirow{4}*{$0.65\times$} \\
    \cmark & \xmark & 65 & 81.5 & \\
    \xmark & \cmark & 65 & 80.4 & \\
    \cmark & \cmark & 65 & {81.7$_{\color{ForestGreen} -0.0}$} & & & \\
  \end{tabular}
  \caption{Ablation studies on the proposed method. 
  Experiments that mark the furthest sampling as \xmark~use random sampling 
  to throw tokens.
  All experiments are conducted by pre-training ViT-B/16~\cite{dosovitskiy2020vit}
  on ImageNet-1K~\cite{russakovsky2015imagenet} for 100 epochs. 
  We count the maximum memory usage and floating-point operations per second~(FLOPs)  
  according to a training iteration of a $224\times 224$ image 
  and report the values relative to the baseline method.
  During pre-training, the progressive reconstruction scheme only increases $7.3\times 10^{-3}$ GFLOPs.}
  \label{tab:ablation}
\end{table*}

Suppose there are $N_m=N\cdot \rho_e$ masked tokens, 
and we throw $N_t=N\cdot \rho_d$ tokens among them. 
For illustration, 
we define $\mathbf{D}\in \mathbb{R}^{N_m\times N_m}$ as the distance matrix, 
where $\mathbf{D}_{ij}$ is the distance between the $i$-th and $j$-th 
masked tokens. 
In this work, 
we use the Euclidean metric to measure distances because it 
requires minimal computational costs. 
The result of token-throwing is denoted as $\mathbf{s} \in \mathbb{R}^{N_m}$.  
$\mathbf{s}_i=1$ means the $i$-th masked token is retained, and $\mathbf{s}_i=0$ means it is thrown. 
The furthest sampling finds the solution as follows:  
\begin{alignat} {2}
  \label{eq:furthest_sampling}
  \mathop{\max}_{\mathbf{s}} \quad & \sum_{i=0}^{N_m-1}\sum_{j=0}^{N_m-1} \mathbbm{1}(s_i=1 \land s_j=1) \cdot \mathbf{D}_{ij}, \\
  \mbox{s.t.}\quad &\sum_{i=0}^{N_m} \mathbf{s}_i=N_m - N_t,& \\
  &\mathbf{s}_i \in \left\{0,1\right\} \forall i \in \left[0, N_m-1\right],
\end{alignat}
where $\mathbbm{1}$ is the indicator function that outputs 1 if ($s_i=1 \land s_j=1$) and 0 otherwise. 
This process is integrated into the 
multi-threaded data preprocessing stage of deep learning frameworks, 
ensuring it does not affect the training speed. 
In implementation, 
we use a greedy strategy to find an 
approximate solution of Eq.~\eqref{eq:furthest_sampling}. 
Please refer to the supplementary material for more details.

\section{Experiments}
\label{sec:experiments}

\subsection{Experiment settings}

In this work, 
we apply our proposed method 
to different frameworks~\cite{he2021masked,Xie_2022_CVPR,wang2023masked,Liu_2023_ICCV,tec,huang2022green} 
to verify its effectiveness and 
follow the settings of corresponding official papers. 
For a fair comparison, 
we pre-train and fine-tune the models on ImageNet-1K~\cite{russakovsky2015imagenet} 
that comprises 1.2 million images and 1,000 categories. 
Following existing works~\cite{he2021masked,yi2022masked}, 
we mainly use the fully fine-tuning protocol for evaluation. 
Specifically, 
we fine-tune ViT-B/16 and ViT-L/16 for 100 and 50 epochs, respectively. 
Please refer to the supplementary material for more details. 

\begin{table}[t]
  \centering
  \setlength{\tabcolsep}{0.5mm}
  \begin{tabular}{lccclcc} 
    & Architecture & Epochs & $\rho_d$ & Top-1 & Flops & Memory \\ \midrule
    MAE & ViT-B/16 & 100 & 0 & 81.7 & $1.00\times$ & $1.00\times$ \\
    MAE & ViT-B/16 & 100 & 50 & 81.1 & $0.72\times$ & $0.64\times$ \\
    +\nameofmethod & ViT-B/16 & 100 & 50 & 81.8 & $0.72\times$ & $0.64\times$ \\
    \midrule
    MAE & ViT-B/16 & 300 & 0 & 82.7 & $1.00\times$ & $1.00\times$ \\
    MAE & ViT-B/16 & 300 & 50 & 82.3 & $0.72\times$ & $0.64\times$ \\
    +\nameofmethod & ViT-B/16 & 300 & 50 & 82.7 & $0.72\times$ & $0.64\times$ \\
    \midrule
    MAE & ViT-B/16 & 800 & 0 & 83.3 & $1.00\times$ & $1.00\times$ \\
    MAE & ViT-B/16 & 800 & 50 & 83.0 & $0.72\times$ & $0.64\times$ \\
    +\nameofmethod & ViT-B/16 & 800 & 50 & 83.3 &$0.72\times$ & $0.64\times$ \\
    \midrule
    MAE & ViT-L/16 & 100 & 0 & 83.3 & $1.00\times$ & $1.00\times$ \\
    MAE & ViT-L/16 & 100 & 50 & 82.4 & $0.87\times$ & $0.82\times$ \\
    +\nameofmethod & ViT-L/16 & 100 & 50 & 83.2 & $0.87\times$ & $0.82\times$ \\
    \midrule
    MAE & ViT-L/16 & 800 & 0 & 85.4 & $1.00\times$ & $1.00\times$ \\
    +\nameofmethod & ViT-L/16 & 800 & 50 & 85.4 & $0.87\times$ & $0.82\times$ \\
  \end{tabular}
  \caption{Training schedules and model scaling. PR-MIM means our method.}
  \label{tab:training_schedules}
\end{table}

\subsection{Experiment results}

\tabref{tab:ablation} examines the effects of the proposed progressive reconstruction scheme and furthest sampling
across different throwing ratios $\rho_d$. 
For example, 
the baseline with a token-throwing ratio $\rho_d$ of 50\%~(\ie, the decoder only operates half of the tokens) 
significantly reduces the computational costs and accelerates the pre-training 
but degrades the Top-1 accuracy from 81.7\% to 81.1\%. 
In contrast, 
our proposed method can mitigate  
this degradation without increasing computational costs, 
apart from the negligible $7.3\times 10^{-3}$ GFLOPs in spatial aggregation. 
The experiments of different throwing ratios also share a consistent trend, 
verifying that our method can learn high-quality representations and 
significantly save costs. 

\myPara{Training schedules and model scaling.}
\tabref{tab:training_schedules} shows 
that the proposed method can adapt to 
longer schedules and larger models. 
For example, 
when pre-training ViT-B/16 for 800 epochs, 
our method 
achieves performance consistent with the standard MAE, 
while \nameofbase~reduces the performance by 0.3\%. 
We also observe that larger models are more susceptible to the negative effects of \nameofbase~in \tabref{tab:training_schedules} and \tabref{tab:arch}. 
When pre-training for 100 epochs, 
\nameofbase~degrades the performance by 0.6\% and 0.9\% with ViT-B/16 and ViT-L/16, respectively. 
But our method mitigates the degradation. 

In this work, 
we mainly focus on the plain vision transformer because it 
is popular in vision representation learning and 
needs to save costs more than other architectures due to 
its high-resolution output. 
\tabref{tab:arch} also shows that our method can apply to other architectures, 
such as Swin transformer~\cite{liu2021Swin}. 
For instance, 
\nameofbase~degrades the performance of Swin-B and Swin-L by 0.3\% and 0.7\% Top-1 accuracy, respectively, 
and our method can eliminate this. 
Meanwhile, 
these results further confirm that 
larger models may be more susceptible to the negative effects of \nameofbase. 

\begin{table}[t]
  \centering
  \setlength{\tabcolsep}{0.5mm}
  \begin{tabular}{lcccclcc} 
    & Target & Epochs & $\rho_d$ & Top-1 & FLOPs & Memory \\ 
    \midrule
    SimMIM~\cite{Xie_2022_CVPR} & \thRows{RGB} & \thRows{100} & $0$ & 81.6 & $1.00\times$ & $1.00\times$ \\
    SimMIM~\cite{Xie_2022_CVPR} & & & $30$ & 81.0 & $0.50\times$ & $0.61\times$ \\
    +\nameofmethod & & & $30$ & 81.7 & $0.50\times$ & $0.61\times$ \\
    \midrule
    SimMIM~\cite{Xie_2022_CVPR} & \tRows{RGB} & \tRows{800} & $0$ & 83.8 & $1.00\times$ & $1.00\times$ \\
    +\nameofmethod & & & $30$ & 83.8 & $0.50\times$ & $0.61\times$ \\
    \midrule
    LocalMIM~\cite{wang2023masked} & \thRows{HOG} & \thRows{100} & $0$ & 83.3 & $1.00\times$ & $1.00\times$ \\
    LocalMIM~\cite{wang2023masked} & & & $65$ & 82.9 & $0.91\times$ & $0.84\times$ \\
    +\nameofmethod & & & $65$ & 83.3 & $0.91\times$ & $0.84\times$ \\
    \midrule
    MFF~\cite{Liu_2023_ICCV} & \thRows{RGB} & \thRows{800} & $0$ & 83.6 & $1.00\times$ & $1.00\times$ \\
    MFF~\cite{Liu_2023_ICCV} & & & $50$ & 83.3 & $0.73\times$ & $0.64\times$ \\
    +\nameofmethod & & & $50$ & 83.8 & $0.73\times$ & $0.64\times$ \\
    \midrule
    TEC~\cite{tec} & \thRows{iBOT} & \thRows{300} & $0$ & 84.8 & $1.00\times$ & $1.00\times$ \\
    TEC~\cite{tec} & & & $50$ & 84.5 & $0.82\times$ & $0.81\times$ \\
    +\nameofmethod & & & $50$ & 84.7 & $0.82\times$ & $0.81\times$ \\
  \end{tabular}
  \caption{Compatibility with different methods. 
  Besides MAE~\cite{he2021masked} in \tabref{tab:ablation}, 
  our method is compatible with different methods.}
  \label{tab:framework}
\end{table}

\begin{table}[t]
  \centering
  \setlength{\tabcolsep}{1.5mm}
  \begin{tabular}{llccc} 
    & Architecture & Epochs & $\rho_d$ & Top-1 \\ 
    \midrule
    GreenMIM~\cite{huang2022green} & \thRows{Swin-B~\cite{liu2021Swin}} & \thRows{100} & 0 & 83.2 \\
    GreenMIM~\cite{huang2022green} & & & 50 & 82.9 \\
    +\nameofmethod & & & 50 & 83.3 \\
    \midrule
    GreenMIM~\cite{huang2022green} & \thRows{Swin-L~\cite{liu2021Swin}} & \thRows{100} & 0 & 83.8 \\
    GreenMIM~\cite{huang2022green} & & & 50 & 83.1 \\
    +\nameofmethod & & & 50 & 83.9 \\
  \end{tabular}
  \caption{Compatibility with Swin transformer, where the masking ratio is 0.75.}
  \label{tab:arch}
\end{table}

\myPara{Generalizing to diverse frameworks.}
Apart from MAE, 
the proposed method can be applied 
to other frameworks, as shown in \tabref{tab:framework}. 
We apply our method to MFF~\cite{Liu_2023_ICCV}, GreenMIM~\cite{huang2022green}, 
TEC~\cite{tec}, and LocalMIM~\cite{wang2023masked} 
and consistently guarantee the representation quality while reducing the costs. 
Regarding MFF, 
our method 
even improves the Top-1 accuracy by 0.2\% 
when accelerating the training. 

SimMIM~\cite{Xie_2022_CVPR} implements masked image modeling differently from MAE. 
It accomplishes the representation extraction and token reconstruction 
in a unified encoder. 
To apply our method to SimMIM, 
we throw a subset of masked tokens to reduce the sequence length of the encoder and 
append the proposed spatial aggregation module after the encoder 
to reconstruct the thrown tokens. 
In \tabref{tab:framework}, 
we reduce the costs of SimMIM 
by 50\% GFLOPs and 39\% memory usagec
without degrading the performance, 
demonstrating that 
our method is orthogonal to different methods.

\begin{table}[t]
  \centering
  \setlength{\tabcolsep}{1.0mm}
  \begin{tabular}{lccclcc} 
    & Architecture & Epochs & $\rho_d$ & Top-1 \\ 
    \midrule
    MAE & \multirow{3}*{ViT-B/16} & 400 & 40 & 82.6$^\dag$ \\
    MAE+AMT~\cite{goodhelp} & & 400 & 40 & 82.8 \\
    MAE+\nameofmethod & & 400 & 50 & 83.0 \\
    \midrule
    SimMIM & \multirow{3}*{ViT-B/16} & 100 & 24 & 81.6 \\
    SimMIM+AMT~\cite{goodhelp} & & 200 & 24 & 80.7 \\
    SimMIM+\nameofmethod & & 100 & 30 & 81.7 \\
  \end{tabular}
  \caption{Comparison with existing partial reconstruction methods. 
  $^\dag$ means that the result is borrowed from~\cite{goodhelp}.}
  \label{tab:other_dual_masking}
\end{table}

\begin{figure}[t]
  \hspace{5pt}
  \begin{overpic}[width=0.97\columnwidth]{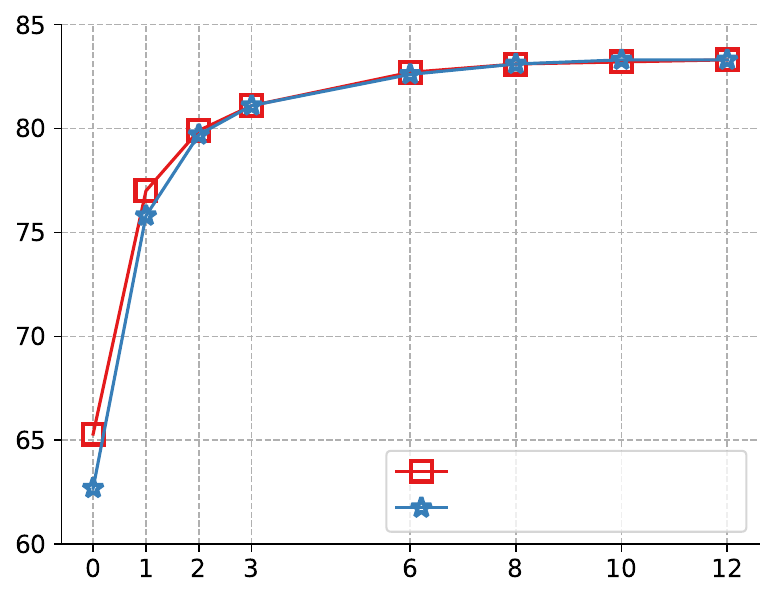}
    \put(36, -2){\small fine-tuning blocks}
    \put(-3, 28){\rotatebox{90}{\small Top-1 accuracy}}
    \put(59, 15.2){\small MAE}
    \put(59, 10.5){\small \nameofmethod{}}
  \end{overpic}
  \caption{Partial fine-tuning results of ViT-B pre-trained for 800 epochs. 
  Fine-tuning 0 and 12 blocks means the linear probing and 
  fully fine-tuning in \tabref{tab:training_schedules}, respectively.}
  \label{fig:part_finetune}
\end{figure}
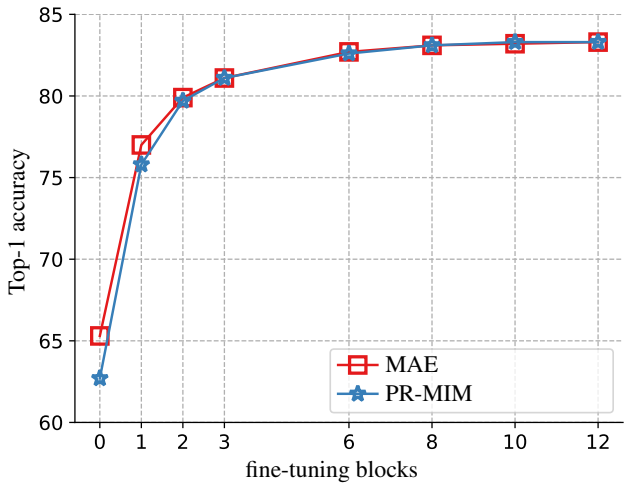

\myPara{Comparison with \nameofbase~methods.} 
\tabref{tab:other_dual_masking} 
compares our method with previous \nameofbase~methods. 
AMT~\cite{goodhelp} utilizes attention-driven masking and throwing to 
mitigate the negative impact of \nameofbase~on performance, 
but attention maps incur extra computational and storage costs. 
Moreover, 
it cannot solve the performance degradation when applying it to SimMIM. 
Instead, 
our simple and effective method 
outperforms AMT using even a higher throwing ratio. 

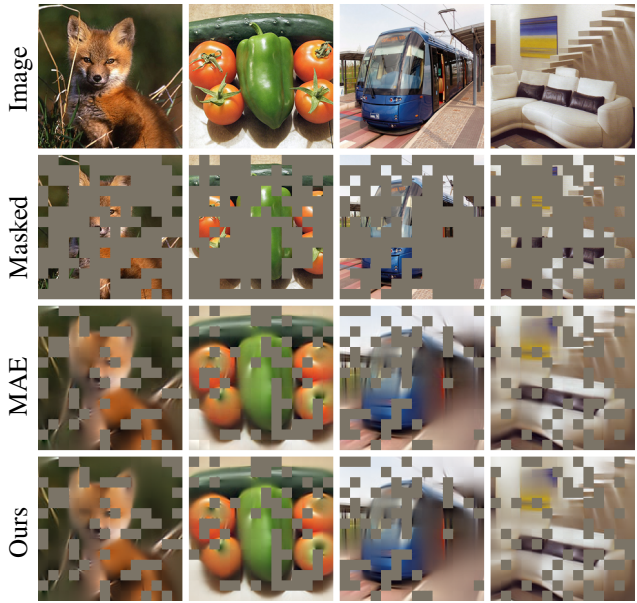
\begin{figure}[t]
  \hspace{8pt}
  \begin{overpic}[width=0.95\columnwidth]{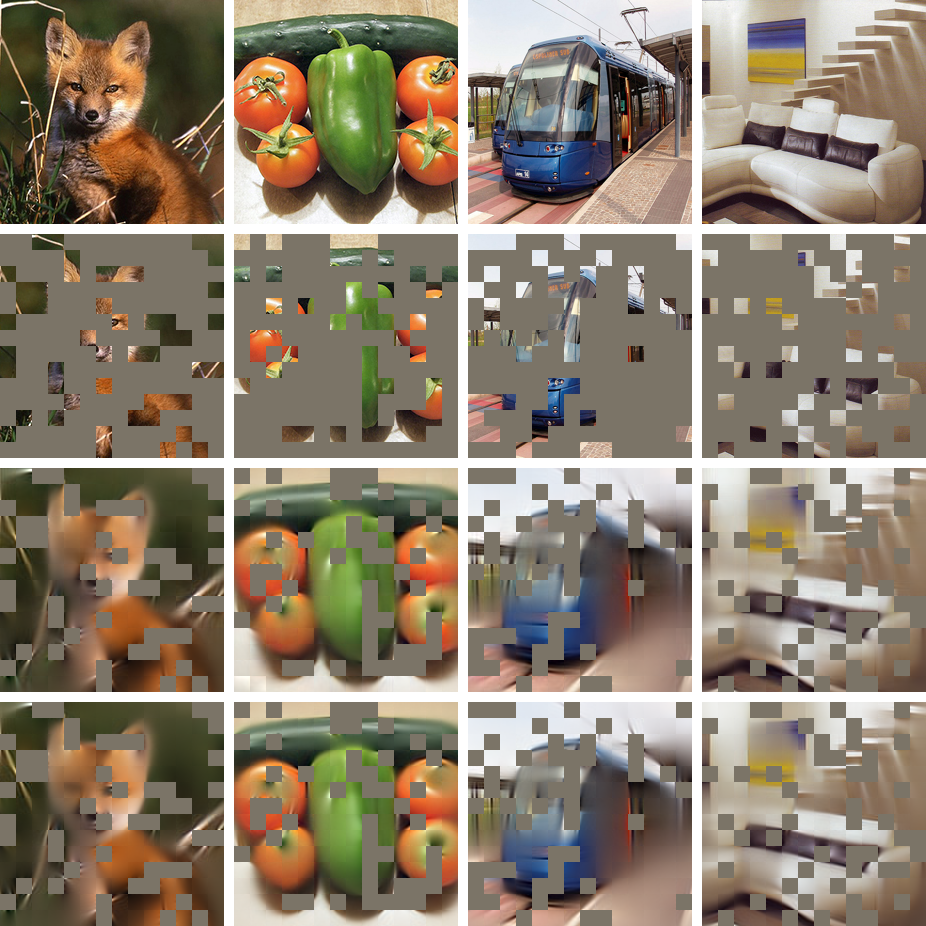}
    \put(-5, 82){\rotatebox{90}{Image}}
    \put(-5, 54){\rotatebox{90}{Masked}}
    \put(-5, 31){\rotatebox{90}{MAE}}
    \put(-5, 7){\rotatebox{90}{Ours}}
  \end{overpic}
  \caption{Reconstruction visualization by different methods. 
  The second row shows the unmasked pixels of the original images, 
  and the third row shows the reconstruction results of masked pixels.}
  \label{fig:visualization_mae}
\end{figure}

\begin{table}[t]
  \centering
  \setlength{\tabcolsep}{1.0mm}
  \begin{tabular}{lcccccccccc} 
    & Epochs & $\rho_d$ & IN-C$\downarrow$ & IN-A & IN-R & IN-S \\
    \midrule
    MAE & 800 & $0$ & 51.2 & 34.6 & 49.4 & 35.2 \\
    MAE & 800 & $50$ & 51.9 & 32.9 & 49.4 & 36.0 \\
    MAE + \nameofmethod & 800 & $50$ & {51.4} & {35.4} & {49.7} & {36.4} \\
    \midrule
    MFF & 800 & $0$ & 49.0 & 37.2 & 51.0 & 36.8 \\
    MFF & 800 & $50$ & 50.5 & 35.6 & 51.7 & 37.6 \\
    MFF + \nameofmethod & 800 & $50$ & 49.9 & 37.9 & 52.1 & 37.6 \\
  \end{tabular}
  \caption{Robust evaluation. We report Top-1 accuracy apart from IN-C~\cite{hendrycks2019robustness}, 
  in which we report mean corruption error.}
  \label{tab:robust}
\end{table}

\myPara{Partial fine-tuning.}
In this work, we mainly adopt 
the evaluation protocol 
of fully fine-tuning models 
because this protocol is popular in MIM. 
Experiments show the proposed method can eliminate the performance degradation caused by \nameofbase~when 
using this protocol. 
Linear probing is another evaluation protocol. 
However, 
pure MIM methods 
aim at learning 
powerful nonlinear representations for downstream tasks 
instead of linearly separable representations. 
Thus linear probing is inapplicable to evaluate them 
as analyzed in~\cite{he2021masked}. 
Following existing works~\cite{he2021masked,yi2022masked}, 
we also evaluate the models by partial fine-tuning. 
As shown in \figref{fig:part_finetune}, 
our method does not fully bridge the degradation in linear probing 
because it does not optimize the linear discriminability of the representations. 
However, fine-tuning only 2 blocks can eliminate 
the performance gap. 
These results, 
along with the results in \secref{sec:downstream}, 
show that 
\nameofmethod{} ensures strong nonlinear representations that are suitable for downstream tasks 
when accelerating training. 

\myPara{Reconstruction visualization.} 
\figref{fig:visualization_mae} 
shows the reconstruction results when using our method. 
Compared to standard MAE, 
we achieve comparable reconstruction quality, 
showing the effectiveness of the extremely lightweight spatial aggregation module.

\begin{table}[t]
  \centering
  \setlength{\tabcolsep}{2.5mm}
  \begin{tabular}{lcccccccccc} 
    & Epochs & $\rho_d$ & iNat$_{18}$ & iNat$_{19}$ \\
    \midrule
    MAE & 800 & $0$ & 72.5 & 78.9 \\
    MAE & 800 & $50$ & 71.8 & 78.2 \\
    MAE + \nameofmethod & 800 & $50$ & {72.4} & {79.0} \\
    \midrule
    MFF & 800 & $0$ & 73.5 & 79.8 \\
    MFF & 800 & $50$ & 72.5 & 78.5 \\
    MFF + \nameofmethod & 800 & $50$ & 73.3 & 79.8 \\
  \end{tabular}
  \caption{Top-1 accuracy of image classification where 
  iNat$_{18}$ and iNat$_{19}$ mean iNaturalist$_{2018}$ and iNaturalist$_{2019}$, respectively.}
  \label{tab:classification}
\end{table}

\subsection{Downstream tasks}
\label{sec:downstream}

We transfer the models pre-trained for 800 epochs on various downstream tasks. 
The results of baseline MAE 
are slightly different from those reported in the official papers~\cite{he2021masked,li2022exploring} 
because the official results use models pre-trained for 1600 epochs. 
We follow the official implementation to fine-tune models for a fair comparison. 

\myPara{Robustness.}
In \tabref{tab:robust}, 
we evaluate the models, which have been fine-tuned on clean ImageNet-1K, 
on four corrupted datasets,  
\ie, ImageNet-A~\cite{hendrycks2021nae}, ImageNet-Renditions~\cite{hendrycks2021many}, 
ImageNet-Sketch~\cite{wang2019learning}, and ImageNet-C~\cite{hendrycks2019robustness}. 
Compared to MAE and MFF, 
\nameofbase~negatively impacts the robustness of ImageNet-A and ImageNet-C. 
Our method mitigates these robustness issues and 
even outperforms the MAE and MFF on ImageNet-A. 
On ImageNet-R and ImageNet-S, 
\nameofbase~does not degrade the robustness, 
and our method brings further enhancement. 

\myPara{Transferring on classification.}
In \tabref{tab:classification}, 
we transfer the models to 
other classification datasets with different domains from ImageNet-1K, 
including iNaturalist$_{2018}$ and iNaturalist$_{2019}$ datasets~\cite{Horn_2018_CVPR}. 
Most hyperparameters follow the settings used for ImageNet-1K, 
except the number of fine-tuning epochs set to 360. 
Compared to MAE, 
\nameofbase~degrades the Top-1 accuracy 
by 0.7\% and 0.7\% on iNaturalist$_{2018}$ and iNaturalist$_{2019}$, respectively. 
Instead, 
our method 
eliminates the degradation and 
achieves performance comparable to the MAE. 
There is also a similar trend in comparison with MFF. 

\begin{table}[t]
  \centering
  \setlength{\tabcolsep}{2.5mm}
  \begin{tabular}{lcccccccccc} 
    & Epochs & $\rho_d$ &  mIoU & mAcc \\
    \midrule
    MAE & 800 & $0$ & 46.8 & 57.6 \\
    MAE & 800 & $50$ & 45.5 & 56.6 \\
    MAE + \nameofmethod & 800 & $50$ & {46.9} & 58.2 \\
  \end{tabular}
  \caption{Semantic segmentation on ADE20K.}
  \label{tab:ade20k}
\end{table}

\myPara{Transferring on semantic segmentation.}
Beyond image-level tasks, 
we also evaluate the pre-trained models on dense-level tasks like semantic segmentation. 
We fine-tune the models on ADE20K~\cite{Zhou_2017_CVPR} for 160,000 iterations. 
Regarding hyperparameters, 
we follow the settings in MAE 
except for the layer-wise learning rate decay, 
which we set to 0.75 instead of 0.65, 
as experiments show that a larger layer-wise decay 
is more suitable for models pre-trained for 800 epochs. 
For a fair comparison, 
we also fine-tune the standard MAE with a decay of 0.75 
and achieve a mIoU of 46.8\%, higher than the 46.1\% reported using original settings~\cite{Liu_2023_ICCV}. 
However, 
\nameofbase~degrades the mIoU by 1.3\% 
when throwing 50\% of the masked tokens. 
Our method avoids this issue 
while keeping the acceleration effect of \nameofbase.

\myPara{Transferring on instance segmentation.}
Following ViTDet~\cite{li2022exploring}, 
we fine-tune Mask RCNN~\cite{He_2017_ICCV} for 100 epochs 
with a simple feature pyramid. 
Meanwhile, we use a batch size of 32 instead of 64 
due to the limitation of computational sources, 
and the learning rate is also reduced by half 
following the linear scaling rule~\cite{goyal2018accurate}. 
As shown in \tabref{tab:coco}, 
we report segmentation mask AP$^{\rm m}$ and 
bounding box AP$^{\rm b}$. 
\nameofbaseb~decreases 
the $\rm AP^{b}$ and $\rm AP^{m}$ by 0.8\% and 0.6\%, respectively. 
Instead, our method can mitigate such degradation 
and achieve performance considerable to the standard MAE. 
These results on downstream tasks show that 
our method can 
maintain the quality of representations 
in different downstream tasks 
while significantly reducing pre-training costs. 

\begin{table}[t]
  \centering
  \setlength{\tabcolsep}{2.5mm}
  \begin{tabular}{lcccccccccc} 
    & Epochs & $\rho_d$ & AP$^{\rm b}$ & AP$^{\rm m}$ \\
    \midrule
    MAE & 800 & $0$ & 51.0 & 45.2 \\
    MAE & 800 & $50$ & 50.2 & 44.6 \\
    MAE + \nameofmethod & 800 & $50$ & {50.8} & {45.2} \\
  \end{tabular}
  \caption{Object detection and instance segmentation segmentation on COCO.}
  \label{tab:coco}
\end{table}

\begin{table}[t]
  \centering
  \setlength{\tabcolsep}{2.0mm}
  \begin{tabular}{lccccc} 
    Kernel Size & $3\times3$ & $5\times5$ & $7\times7$ & $9\times9$ & $11\times11$ \\
    \midrule
    Top-1 & 81.7 & 81.7 & 81.8 & 81.7 & 81.6 \\
  \end{tabular}
  \caption{Ablation studies on the kernel sizes in the progressive reconstruction. 
  The ViT-B/16 models are pre-trained on ImageNet-1K for 100 epochs.}
  \label{tab:kernel_size}
\end{table}

\subsection{Ablation studies}

\myPara{Components.} 
In \tabref{tab:ablation}, 
we perform ablation studies on the progressive reconstruction scheme and furthest sampling 
to analyze their impact at different throwing ratios $\rho_d$. 
Without them, 
performance drops as $\rho_d$ increases. 
Across different throwing ratios, 
the progressive reconstruction consistently mitigates 
the performance degradation caused by \nameofbase, 
and the furthest sampling also outperforms random sampling.
Finally, combining them yields the best results 
to achieve efficient trade-offs in memory, computation, and performance. 

\myPara{Spatial receptive field.} 
\tabref{tab:kernel_size} 
evaluates the impact of different kernel sizes for spatial aggregation 
within the progressive reconstruction scheme.
Results indicate that while kernel sizes 
$3\times3$ and 
$5\times5$ both yield 81.7\% Top-1 accuracy, an increase to 
$7\times7$ slightly improves performance to 81.8\%. 
However, further increasing the kernel size to 
$9\times9$ and 
$11\times11$ does not yield additional gains, with performance slightly decreasing to 81.6\%.
These results suggest that a moderate local receptive field size, such as 
$7\times7$, is enough for effective reconstruction.

\myPara{Design of spatial aggregation.}
We evaluate different designs of spatial aggregation in \tabref{tab:decoder_design}, 
including depth-wise convolution, Transformer block, and ConvNeXt block. 
Among them, 
the simple depth-wise convolution achieves high accuracy (81.8\% Top-1) with minimal cost (only 
$7.3\cdot 10^{-3}$ GFLOPs), 
making it both efficient and effective to reconstruct thrown tokens. 
Although Transformer and ConvNeXt blocks also yield strong results, 
they significantly increase FLOPs (0.65G and 0.41G, respectively). 
Additionally, 
we try average pooling, a simpler operation, 
but the model fails to converge due to its inability to model spatial dependencies. 
Thus, depth-wise convolution is chosen for balancing accuracy and efficiency.

\begin{table}[t]
  \centering
  \setlength{\tabcolsep}{3.0mm}
  \begin{tabular}{lcc} 
    & Top-1 & FLOPs \\ 
    \midrule
    Average pooling & fail & - \\
    Depth-wise convolution & 81.8 & $7.3\cdot 10^{-3}$G \\
    Transformer block & 81.8 & 0.65G \\
    ConvNeXt block & 81.7 & 0.41G \\
  \end{tabular}
  \caption{Different designs of the spatial aggregation module in progressive reconstruction.}
  \label{tab:decoder_design}
\end{table}

\section{Conclusion}
\label{sec:conclusions}

In this work, 
we explore integrating \nameofbase, 
a technique that reconstructs only a subset of masked tokens and throws the other ones, 
into 
masked image modeling to save pre-training costs 
without compromising the quality of representations. 
In response to the issue that \nameofbase~does not utilize 
each token in pre-training, 
our proposed progressive reconstruction scheme and 
furthest sampling 
effectively reconstruct 
the tokens thrown by \nameofbase, 
ensuring that all tokens are adequately involved in supervision. 
Importantly, 
the proposed method increases minimal computational costs and 
thus preserves the acceleration benefits of \nameofbase~while 
mitigating the associated performance degradation. 
Extensive experiments 
across 
different downstream tasks demonstrate the effectiveness of our method.

{
    \small
    \bibliographystyle{ieeenat_fullname}
    \bibliography{main}
}

\clearpage
\setcounter{page}{1}
\maketitlesupplementary

\RestyleAlgo{ruled}
\SetKwComment{Comment}{/* }{ */}
\begin{algorithm}
  \caption{Furthest sampling}\label{alg:implementation_furthest_sampling}
  \KwData{$\mathbf{D} \in \mathbb{R}^{N_m\times N_m}$, $N_m$, and $N_t$}
  \KwResult{$s \in \mathbb{R}^{N_m}$}
  $s \gets {\left\{0\right\}}_{N_m}$ \Comment*[r]{Initialization}
  $tokens \gets {\rm empty\_set}()$\; ~\\
  $i \gets {\rm randomint}(0, N_m-1)$ \Comment*[r]{Choose the first retained token}
  $s_{i} \gets 1$\;
  $tokens.{\rm append(i)}$\; ~\\
  \For{$k \leftarrow 1$ \KwTo $N_m-N_t-1$}{
    $\hat{\mathbf{D}} \gets \mathbf{D}_{[:,tokens]} \in \mathbb{R}^{N_m\times k}$ \Comment*[r]{Columns corresponding to retained tokens} ~\\
    $i\gets \mathop{\arg\max}_{i} \left\{\min(\hat{\mathbf{D}}, {\rm axis}=1)\right\}$ \Comment*[r]{The $i$-th token is retained} ~\\
    $s_{i} \gets 1$\;
  $tokens.{\rm append(i)}$\;
  }
\end{algorithm}

\section{Implementation of furthest sampling}
\label{sec:implementation_furthest_sampling}

Given $N_m$ masked tokens, 
the furthest sampling throws $N_t$ tokens and retains $N_m - N_t$ tokens. 
In implementation, 
we use a greedy strategy to find the approximate solution of Eq.~(1) of the manuscript, 
as shown in \algref{alg:implementation_furthest_sampling}. 
After the first retained token is chosen by random sampling, 
this process iteratively 
retains a new token that is furthest from the already retained tokens, 
ensuring a dispersed selection of retained tokens. 

\begin{table}[t]
  \centering
  \setlength{\tabcolsep}{0.3mm}
  \caption{\textbf{Pre-training details.}}
  \begin{tabular}{lcccccccccc} 
    item & value \\ \toprule
    optimizer & AdamW~($\beta_1=0.9$ and $\beta_2=0.95$) \\
    base learning rate & 1.5e-4 \\
    learning rate & 2.4e-3 \\
    weight decay & 0.05  \\
    batch size & 4096 \\
    warmup epochs & 40 \\
    epochs & 100/300/800 \\
    mask ratio & 0.75 \\
    norm pixel loss & True \\
    learning rate schedule & cosine decay \\
  \end{tabular}
  \label{tab:pre_training_details}
\end{table}

\section{Implementation details}
\label{sec:implementation_details_appendix}

\begin{table}[t]
  \centering
  \setlength{\tabcolsep}{0.3mm}
  \caption{\textbf{Fine-tuning details.}}
  \begin{tabular}{lcccccccccc} 
    item & value \\ \toprule
    optimizer & AdamW~($\beta_1=0.9$ and $\beta_2=0.999$) \\
    base learning rate & 5e-4(B),1e-3(L) \\
    learning rate & 2e-3(B),4e-3(L) \\
    weight decay & 0.05 \\
    batch size & 1024 \\
    warmup epochs & 5 \\
    epochs & 100~(B),50~(L) \\
    layer-wise lr decay & 0.65(B),0.75(L) \\
    drop-path rate & 0.1 (B),0.2(L) \\
    data augmentation & RandAug (9, 0.5) \\
    reprob & 0.25 \\
    mixup & 0.8 \\
    cutmix & 1.0 \\
    label smoothing & 0.1 \\
    learning rate schedule & cosine decay \\
  \end{tabular}
  \label{tab:fine_tuning_details}
\end{table}

\tabref{tab:pre_training_details} and \tabref{tab:fine_tuning_details} 
summarize the  
hyperparameters used for pre-training and fine-tuning, respectively. 
These hyperparameters are used for MAE-based experiments, 
and the other experiments follow the official settings of the corresponding papers. 
For fine-tuning, 
the learning rate and layer-wise learning-rate decay rate 
are slightly different from the official paper of MAE and 
are adjusted to be consistent with 
the official codes. 
This is because the official codes of MAE 
use normalized pixels as the reconstruction targets, but 
the official paper uses raw pixels as the default setting. 
Using normalized pixels achieves a better performance, 
thus we follow the settings of official codes. 

\end{document}